\documentclass[11pt,a4paper]{article}
\usepackage{emnlp2021}
\usepackage{times}
\usepackage{latexsym}

\usepackage{microtype}

\newcommand{\cmark}{\ding{51}}%
\newcommand{\xmark}{\ding{55}}%
\usepackage{array}\newcolumntype{P}[1]{>{\centering\arraybackslash}p{#1}}
\usepackage{makecell}
\usepackage{graphicx}
\usepackage{amsmath}
\usepackage{amssymb}
\usepackage{pifont}
\usepackage{booktabs}
\usepackage{enumitem}
\usepackage{soul}
\usepackage{moresize}
\usepackage{multicol}
\usepackage{multirow}
\usepackage{adjustbox}

\title{Beyond the Tip of the Iceberg: Assessing Coherence of Text Classifiers}

\author{Shane Storks \and Joyce Chai \\
       Computer Science and Engineering Division \\
       University of Michigan \\
        Ann Arbor, MI 48109, USA \\
        \texttt{\{sstorks, chaijy\}@umich.edu} \\  }

\begin{document}
\maketitle
\begin{abstract}

As large-scale, pre-trained language models achieve human-level and superhuman accuracy on existing language understanding tasks, statistical bias in benchmark data and probing studies have recently called into question their true capabilities. For a more informative evaluation than accuracy on text classification tasks can offer, we propose evaluating systems through a novel measure of prediction coherence. We apply our framework to two existing language understanding benchmarks with different properties to demonstrate its versatility. Our experimental results show that this evaluation framework, although simple in ideas and implementation, is a quick, effective, and versatile measure to provide insight into the coherence of machines' predictions.

\end{abstract}

\section{Introduction}\label{sec:intro}
Large-scale, pre-trained contextual language representations \cite{devlinBERTPretrainingDeep2018,radfordImprovingLanguageUnderstanding2018a,raffelExploringLimitsTransfer2019,brownLanguageModelsAre2020} have approached or exceeded human performance on many existing language understanding benchmarks. 
However, due to increasing complexity and concerns of statistical bias enabling artificially high performance \cite{schwartzEffectDifferentWriting2017,poliakHypothesisOnlyBaselines2018,nivenProbingNeuralNetwork2019,minSyntacticDataAugmentation2020}, the coherence of these state-of-the-art systems and their alignment to humans is not well understood.

\begin{figure}
  \centering
  \includegraphics[width=0.48\textwidth]{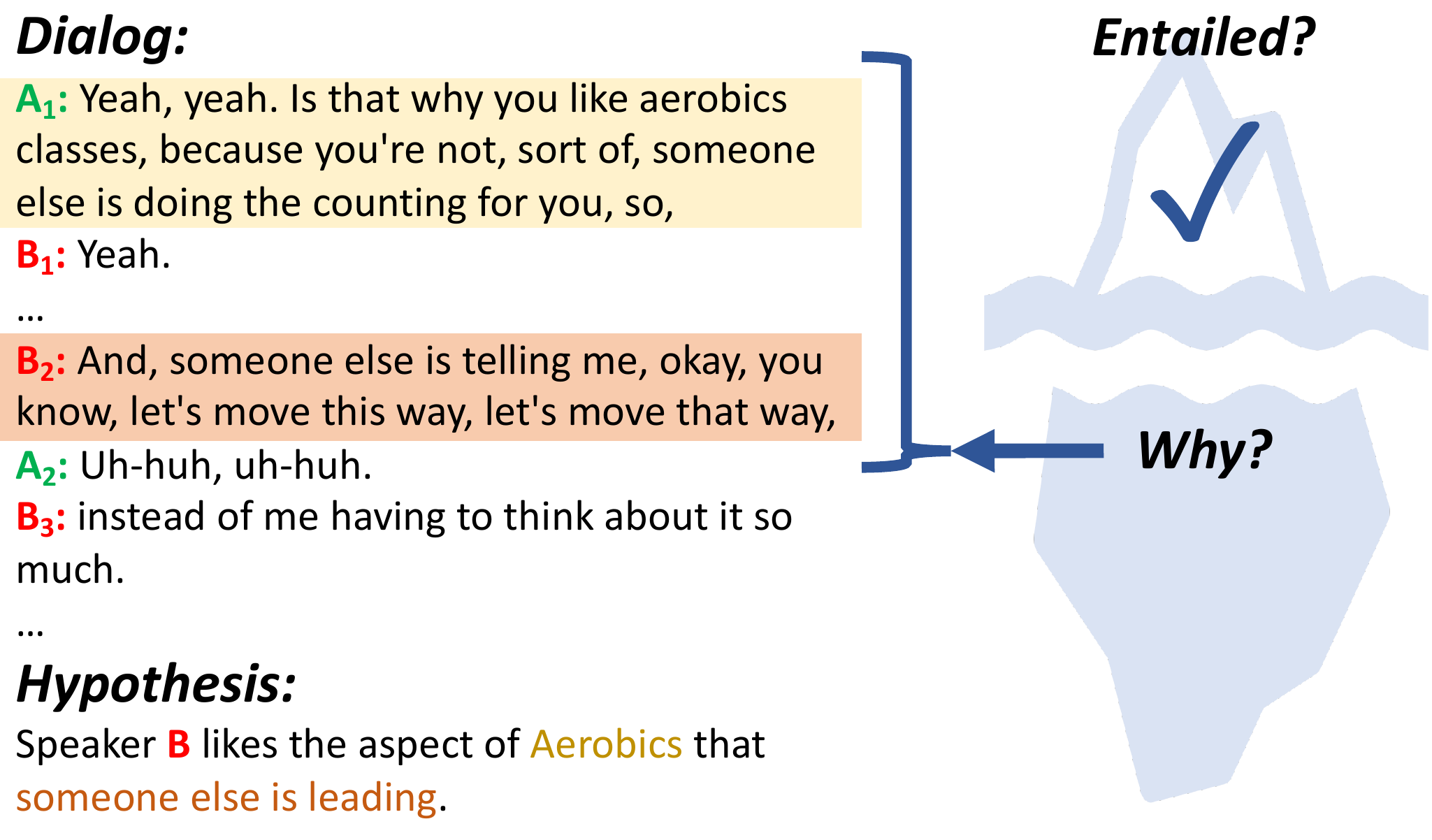}
  
  \caption{In Conversational Entailment~\cite{zhang-chai-2010-towards}, systems only predict whether a hypothesis is entailed by a dialog, while ignoring the underlying evidence in the discourse toward this conclusion.}
  \vspace{-10pt}
  \label{fig:idea1}
\end{figure}

This is perhaps because benchmarks geared toward language understanding only cover the tip of the iceberg, typically focusing on a high-level end task rather than diving deeper into the kind of coherent, robust understanding that takes place in humans. Language understanding in machines is often boiled down to text classification, where a classifier is tasked with recognizing whether a text contains a particular semantic class, e.g., textual entailment~\cite{daganPASCALRecognisingTextual2005,bowmanLargeAnnotatedCorpus2015}, commonsense implausibility~\cite{roemmeleChoicePlausibleAlternatives2011,mostafazadehCorpusClozeEvaluation2016,biskPIQAReasoningPhysical2020}, or combinations of several phenomena meant to serve as comprehensive diagnostics \cite{poliakDiverseNaturalLanguage2018,wangGLUEMultiTaskBenchmark2018,wangSuperGLUEStickierBenchmark2019}. Without regard to the underlying evidence used to reach a conclusion, systems are rewarded for correct predictions on the task without ``showing their work.''

To make meaningful improvement on machine language understanding, it is important to have more informative performance measures. 
To address this issue, the contribution of this paper is to introduce a novel model- and task-agnostic evaluation framework that allows a quick assessment of text classifiers' ability in terms of the coherence of their predictions.
We apply our framework to two existing language understanding benchmarks of different genres to demonstrate its versatility. Our results support recent findings of spurious behaviors in fine-tuned large LMs, and show that our framework, although simple in ideas and implementation, is effective as a quick measure to provide insight into the coherence of machines' predictions.

\section{Related Work}
In the face of data bias and uninterpretability of large LMs, past work has proposed methods to robustly interpret and evaluate them for various tasks and domains. 
Some work has sought to probe contextual language representations through various means to better understand what knowledge they hold and their correspondence to syntactic and semantic patterns \cite{tenneyWhatYouLearn2018,hewittStructuralProbeFinding2019,jawaharWhatDoesBERT2019,tenneyBERTRediscoversClassical2019a}.
Meanwhile, behavior testing approaches have also been applied to understand model capabilities, from automatically removing words in language inputs and examining model performance as the input becomes malformed or insufficient for prediction~\cite{li2016understanding,murdoch2018beyond,hewittStructuralProbeFinding2019}, to curating fine-grained testing data to measure performance on interesting phenomena~\cite{zhou2019going,ribeiroAccuracyBehavioralTesting2020}.
Similar work has used specialized natural language inference tasks~\cite{welleck-etal-2019-dialogue,uppal-etal-2020-two}, logic rules~\cite{li-etal-2019-logic,asai-hajishirzi-2020-logic}, and annotated explanations \cite{deyoung-etal-2020-eraser, jhamtani-clark-2020-learning} to support and evaluate consistency and coherence of inference in these models. Other works have studied coherence of discourse through the proxy task of sentence re-ordering \cite{lapata-2003-probabilistic,Logeswaran_Lee_Radev_2018}.
Different from these previous works that focus only on specific tasks or methods, or require heavy annotation,
this paper introduces an easily-accessed, versatile evaluation of machine coherence from a small amount of additional annotation.

\section{Coherent Text Classification}\label{sec:coherentcr}

For any text classification task requiring reasoning over a discourse, a coherent classifier should use the same evidence as humans do in reaching a conclusion. For any positive example, we expect that there are specific regions of the text which contain the semantic class of interest and thus directly contribute to the positive label. Conversely, for any negative example, there should be no such regions of the text. At a high level, we will propose a coherence measure that captures whether classifiers can give consistent and human-aligned predictions on these regions to support the end task conclusion.

Depending on specific tasks, this measure can have different implementations while maintaining the same high-level goal. 
In the following sections, we will use two example benchmark datasets, Conversational Entailment (\textbf{CE}) from \citet{zhang-chai-2010-towards} and Abductive Reasoning in narrative Text (\textbf{ART}) from \citet{ch2019abductive}, to illustrate how the coherence measure can be applied.  We intentionally chose these two distinctive benchmark datasets for our investigation. CE is formulated as a textual entailment task, while ART is a multiple-choice text plausibility classification task. CE is small-scale, created over ten years ago before the era of deep learning, while ART is a large-scale ($\sim$171k examples) dataset created more recently. Through these two different datasets, we aim to demonstrate the versatility of this framework.

\subsection{Coherence in Textual Entailment}

\begin{figure}
  \centering
  \includegraphics[width=0.48\textwidth]{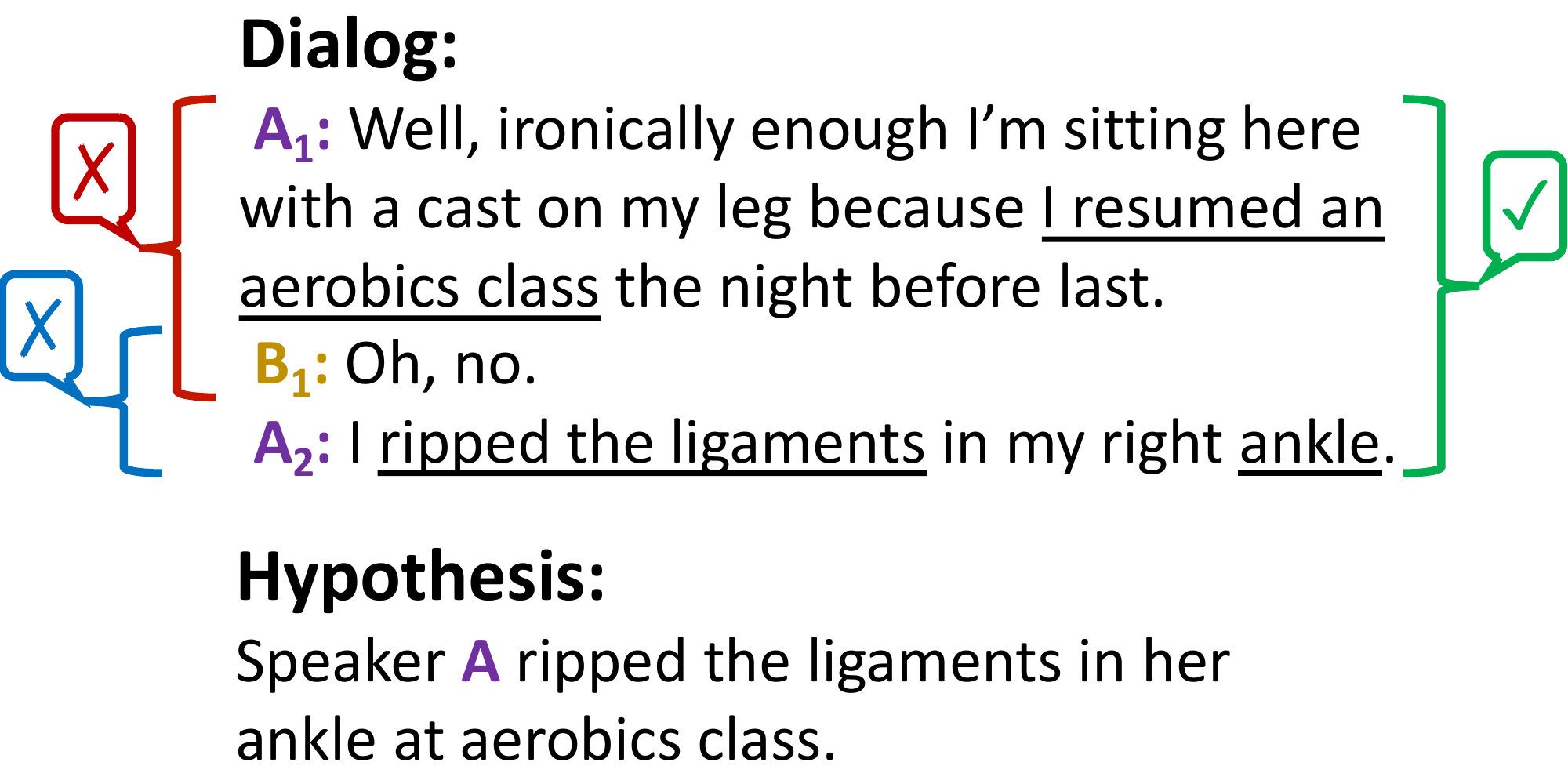}
  \caption{In CE, we label each sub-span of dialog with whether it entails the hypothesis ({\cmark } for yes, {\xmark }  for no).}
    \vspace{-10pt}
  \label{fig:spans}
\end{figure}

CE poses a textual entailment task where context is given as several turns of a natural language dialog, and we must determine whether the dialog entails a hypothesis sentence. All required information is explicitly given in the dialog. In each positive example, only some dialog turns directly contribute to the entailment, while others are irrelevant to the hypothesis.
For example, as shown in Figure~\ref{fig:idea1}, turns $A_1$ and $B_2$ together entail the hypothesis, while others are not necessary for entailment.

As shown in Figure~\ref{fig:spans} for CE, we can label individual spans of a discourse that entails a hypothesis with whether or not consecutive sub-spans of the discourse also entail the hypothesis. Here, while the entire dialog from $A_1$ through $A_2$ entails the hypothesis, the spans from $A_1$ through $B_1$ and $B_1$ through $A_2$ do not, as they omit details required by the hypothesis.
Given an example of length $N$,\footnote{Length can be defined in units of dialog turns, sentences, paragraphs, or other appropriate units of the text. Text should be decomposed such that individual sub-spans are not malformed or fragmented, so token- and character-level sub-spans will typically be inappropriate for this evaluation.} we can decompose it into $N+ {N \choose 2}$ possible consecutive sub-spans\footnote{There are ${N \choose 2}$ combinations of starting and ending points for multi-sentence sub-spans, plus $N$ individual sentences.} to label with human judgements. 

For a correctly classified example, we can then perform inference on all sub-spans. If the system additionally classifies all of them correctly, we consider the prediction to be coherent. We then calculate \textbf{coherence} on the task as the percentage of examples coherently classified. Extremely simple to compute, this provides valuable insight beyond the surface of end task accuracy, measuring how well the classifier's perceived evidence toward the conclusion aligns with that of humans.
Alternatively, the average sub-span accuracy may be considered as a more lenient measure.

\subsection{Coherence in Plausibility Classification}

\begin{figure}
  \centering

    \includegraphics[width=0.45\textwidth]{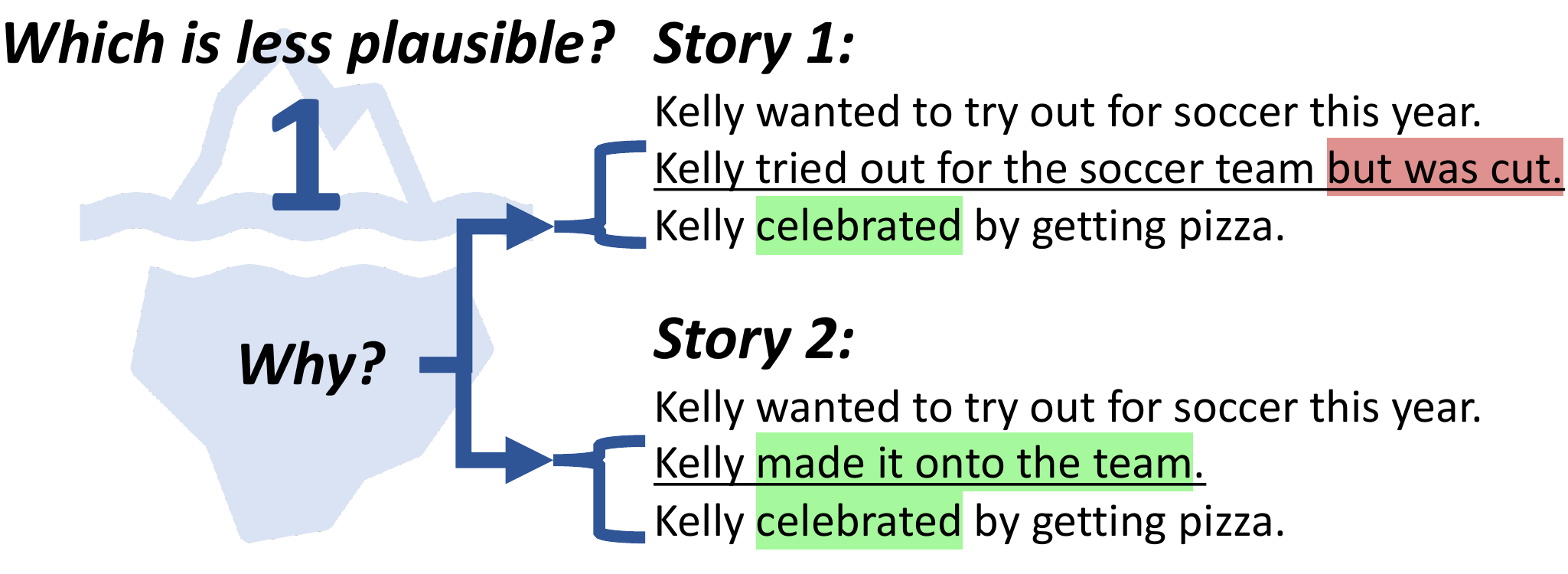}

  \caption{In Abductive Reasoning in narrative Texts~\cite{ch2019abductive}, systems only compare two texts by their commonsense plausibility, ignoring which parts of the stories support this conclusion.}
  \vspace{-10pt}
  \label{fig:idea2}
\end{figure}

ART, meanwhile, is a multiple-choice text classification benchmark for commonsense plausibility recognition. The task is to determine which of two candidate sentences most plausibly fits between two given context sentences when considering commonsense constraints on the world. This translates naturally into a choice between two three-sentence stories (differing only by the second sentence), one of which has some implausibility (the positive choice). For example, as shown in Figure~\ref{fig:idea2}, Story 1 is implausible because while the second sentence describes a negative event, the third sentence indicates celebration. Meanwhile, in Story 2, the agent is celebrating a positive event.

\begin{figure}
  \centering
  \includegraphics[width=0.48\textwidth]{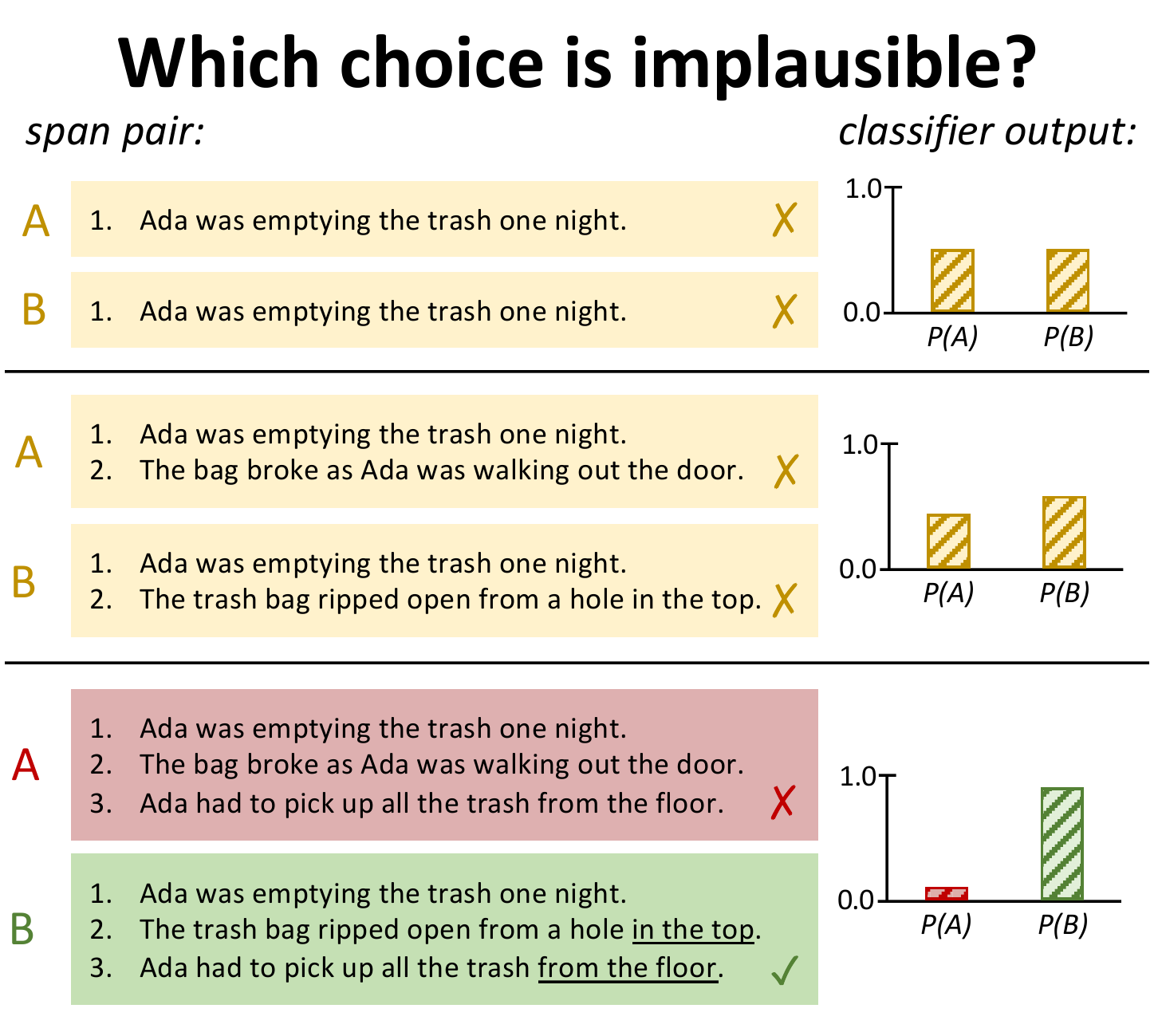}
  \caption{In ART, a multiple-choice text classification problem, we can label sub-spans with the least plausible choice, although in some cases, both choices are plausible. To address this, we consider the classifier's posterior probability for each choice; it is ideal if the classifier has low confidence in such instances. }
  \vspace{-10pt}
  \label{fig:mc}
\end{figure}

\paragraph{Multiple-choice tasks.}
To account for multiple-choice tasks like ART, where we identify one of two texts to be semantically implausible, we must adjust this setup. We still consider sub-spans of the context, breaking down each pair of texts into $N+ {N \choose 2}$ pairs of sub-spans. Intuitively, the model's choice on each pair should again align with that of humans. However, there is a possibility that none of the texts contain the positive class. In such cases, the classifier should not make a confident prediction, and instead believe the texts are equally likely. Confidence should be defined based on the classifier's internal model of the probability distribution over all possible class labels, i.e., text choices (typically calculated by applying softmax over the activations of several neural network branches). This is conceptually visualized in Figure~\ref{fig:mc}, where a classifier should only become confident that Story B is implausible once both the second and third sentence are present, as \textit{the trash} is less likely to end up on \textit{the floor} with a \textit{hole in the top} of the bag.

Generally, let $T_{a:b}$ represent the consecutive sub-sequence of text $T$ from unit $a$ through $b$, e.g., sentences $a$ through $b$ of text $T$.
Consider a set $S_{1:N}$ of $M$ texts of length $N$ such that $S = \{ T^1_{1:N}, T^2_{1:N}, \cdots , T^M_{1:N} \}$, and a classifier $f$ such that $f(S_{1:M}) \in [1, M] $.\footnote{While text choices may be different lengths, this can be trivially resolved by padding.} 
When classifying a set $S_{a:b}$, 
let $f(S_{a:b}) = c^*$ be considered a \textit{confident} prediction if $\underset{c \in [1,c^*) \lor (c^*, M] }{\max} \left( p(c^*) - p(c) \right) \geq \rho$, where $p(c)$ refers to probability of class $c$ under the classifier's output distribution, and $\rho$ is a confidence threshold. Where there is no positive text within $S_{a:b}$, then the desired outcome (ground truth) is for $f(S_{a:b})$ to be a non-confident prediction. This should be reflected in the calculation of coherence.

\section{Coherence of SOTA Classifiers}\label{sec:results}
Using our framework, we next establish baseline measures of coherence on the two benchmarks. The source code and data for our empirical study are shared with the community on GitHub.\footnote{\url{https://github.com/sled-group/Verifiable-Coherent-NLU}}

\subsection{Enabling Coherence Evaluation}
To enable the type of evaluation described in Section~\ref{sec:coherentcr} for our benchmarks, additional annotation is required.
CE contains 50 unique dialog sources from the Switchboard corpus~\cite{switchboard}.
We randomly selected 10 testing sources to form the test set and left all remaining sources for training and validation, creating an 80\%/20\% split for training and validation (703 examples) versus testing (178 examples). We annotated the positive examples in the test set with the range of dialog turns entailing the hypothesis, allowing us to generate ground truth labels for the coherence measurement. Examples were labeled by two separate annotators and cross-verified with a near-perfect Cohen's $\kappa$~\cite{cohen} of 0.91, then a third annotator resolved any disagreements.

To transfer ART to our framework, we annotated 200 random examples from the public validation set (1532 examples) with the evidence for implausibility. There are 3 possible cases in implausible story choices: 1) the second sentence conflicts with the first and/or third sentence, 2) the second sentence is malformed or nonsense, presumably due to annotation error or adversarial filtering~\cite{zellersSWAGLargeScaleAdversarial2018}, and 3) the first and third sentence conflict with each other by default, and the second sentence does not resolve this. These cases are labeled by two annotators then merged with a fair Cohen's $\kappa$ of 0.30 (perhaps lower due to subjectivity of commonsense-based problems), and a third annotator again resolving disagreements. 11 examples were discarded as two annotators agreed that both story choices were entirely plausible, presumably due to annotation error in ART.

\begin{table*}
    \centering

    \normalsize CE, \textit{test}: \footnotesize
    \vspace{1.5pt}
    
    \begin{tabular}{P{1.7cm}|P{2.4cm}|P{5.0cm}|P{5.0cm}}\toprule
            \textbf{Model} &  \textbf{Accuracy} (\%) &
            \textbf{Strict Coherence} ($\Delta$; \%) & \textbf{Lenient Coherence} ($\Delta$; \%) \\\midrule

        {majority} & 57.8 & --  & --   \\\midrule
        \textsc{BERT} & 55.8  & 28.5 $\:\:$ (-27.3) & 35.7 $\:\:$ (-20.1)   \\ 
        \textsc{RoBERTa} & 70.9 & 39.0 $\:\:$ (-31.9) & 47.5 $\:\:$ (-23.4)  \\
        \, $\hookrightarrow$  + \textsc{mnli} & 78.5 & 50.6 $\:\:$ (-27.9) & 58.2 $\:\:$ (-20.3) \\        
        \textsc{DeBERTa} & 67.4 & 37.2 $\:\:$ (-30.2) & 45.2 $\:\:$ (-22.2) \\\bottomrule
        
    \end{tabular}
    
    \vspace{6pt}
    \normalsize ART, \textit{validation}: \footnotesize
    \vspace{1.5pt}
    
    \begin{tabular}{P{1.7cm}|P{2.4cm}|P{4.0cm}|P{0.6cm}|P{4.0cm}|P{0.6cm}}\toprule
            \textbf{Model} &  \textbf{Accuracy} (\%) & \textbf{Strict Coherence} ($\Delta$; \%) & $\rho$ & \textbf{Lenient Coherence} ($\Delta$; \%) & $\rho$ \\\midrule
            
        {majority} & 55.0 $\:\:$ (50.1) & -- & --  & --  \\\midrule
        \textsc{BERT} & 66.7 $\:\:$ (66.7) & 42.3 $\:\:$ (-24.4) & 0.15 & 43.7 (-23.0) & 0.85 \\ 
        \textsc{RoBERTa} & 87.8 $\:\:$ (84.2) & 55.0 $\:\:$ (-32.8) & 0.1 & 59.3 (-28.5) & 0.05 \\
        \textsc{DeBERTa} & 88.4 $\:\:$ (85.7) & 59.8 $\:\:$ (-28.6) & 0.85 & 61.8 (-26.6) & 0.95 \\\midrule

    \end{tabular}\normalsize    
    
    \normalsize
    \caption{Accuracy, strict coherence, and lenient coherence on CE and ART for state-of-the-art text classifiers. $\Delta$ is the total performance drop from the classification accuracy to each coherence measure, and each $\rho$ is the confidence threshold achieving the highest coherence. For ART, accuracy on the full validation set is given in parentheses.}
    \vspace{-10pt}
    \label{tbl:main results}
\end{table*}

\subsection{Empirical Results}
We evaluate three state-of-the-art, transformer-based language models from recent years: \textsc{BERT}~\cite{devlinBERTPretrainingDeep2018}, \textsc{RoBERTa}~\cite{liuRoBERTaRobustlyOptimized2019}, and \textsc{DeBERTa}~\cite{heDeBERTaDecodingenhancedBERT2021}.\footnote{We use the ``large'' configuration of all models, which have 24 hidden layers and 16 attention heads.} On CE, we additionally apply transfer learning from MultiNLI~\cite{williamsBroadCoverageChallengeCorpus2017}, a large-scale textual entailment dataset with some dialog-based problems. We measure both the \textit{accuracy}, i.e., the proportion of instances where the end task prediction is correct, and \textit{coherence} of models on respective evaluation sets. Specifically, we consider two kinds of coherence: strict and lenient. Given a set of evaluation instances, \textit{strict coherence} refers to the proportion of instances where the end task prediction is not only correct, but also coherent as described in Section~\ref{sec:coherentcr}. While strict coherence only rewards systems for examples where all sub-span predictions are correct, \textit{lenient coherence} averages the sub-span accuracy over all examples for a less rigid reward. We include this alternate form of coherence to accommodate some disagreement with our annotations (which can be subjective based on measured inter-annotator agreement) without severe penalty.

\paragraph{Training details.}
Following common practice, systems are trained with cross-entropy loss toward the end task of text classification, maximizing accuracy on the validation set for model selection. 
On CE, we used 8-fold cross-validation split by dialog sources, then re-trained the model with the highest average validation accuracy on all folds.

Pre-trained model parameters and implementations come from HuggingFace \texttt{transformers} \cite{Wolf2019HuggingFacesTS},\footnote{\url{https://huggingface.co/transformers/}} each trained with the AdamW optimizer \cite{loshchilovDecoupledWeightDecay2018}. We performed a grid search over a wide range of learning rates and a maximum of 10 epochs. Training batch sizes are fixed based on available GPU memory. Selected hyperparameters can be found in Appendix~\ref{apx: hyperparam}.

\paragraph{Discussion of results.}
Results on the test set of CE and public validation set of ART are listed in Table~\ref{tbl:main results}. All results show a statistically significant drop in performance from classification accuracy to strict coherence under a McNemar test~\cite{mcnemarNoteSamplingError1947} with $p < \text{1e-5}$, some dropping below majority-class accuracy. While lenient coherence is slightly higher for both tasks, we still see large drops from accuracy. This demonstrates that while our text classifiers can achieve high classification accuracy on CE and ART, they do not deeply understand the tasks. Much of their performance is supported by incoherent intermediate predictions.
Although pre-training on MultiNLI improves the end task accuracy on CE, it still suffers from comparably significant drops to the coherence measures.
On ART, while all models see significant performance drops, \textsc{DeBERTa}, the state-of-the-art system for the task, achieves the best accuracy and coherence measures, as well as the highest chosen $\rho$ values, which generally indicates more confident predictions. Even though it only marginally outperforms \textsc{RoBERTa} in accuracy, we see larger improvements in coherence measures and the chosen $\rho$, suggesting \textsc{DeBERTa} is more robust.

\section{Conclusion}
In this work, we proposed a simple and versatile method to evaluate the coherence of text classifiers, particularly targeting the problem where end task prediction depends on a discourse rather than a single sentence.
\textit{By annotating a small amount of data in a benchmark, this method supports a quick assessment on whether machines' end task performance is supported by coherent intermediate evidence.} Future work driven by benchmarks should consider similar examination beyond the end task accuracy, 
whether this be through our proposed coherence measures or other appropriate means. As we showed, such effort is quite straightforward, and can drive progress toward more powerful classifiers that can support human-aligned reasoning.

\section*{Acknowledgements}
This work was supported in part by IIS-1949634 from the National Science Foundation. We thank Bri Epstein and Haoyi Qiu for their diligent annotation work. We also thank the anonymous reviewers for their helpful comments and suggestions.

\bibliography{references_long}
\bibliographystyle{acl_natbib}

\clearpage

\appendix

\section{Model Training Details}\label{apx: hyperparam}
The selected hyperparameters for each model presented in the paper are listed in Table~\ref{tbl:hyperparams}.

\begin{table}
    \centering
    \footnotesize

    \begin{tabular}{P{0.7cm}|P{2.2cm}|P{0.8cm}|P{1.2cm}|P{0.6cm}}\toprule
            \textbf{Task} &
            \textbf{Model} & \textbf{Batch Size} & \textbf{Learning Rate} & \textbf{Ep.} \\\midrule
            CE & \textsc{BERT} & 32 & 7.5e-6 & 8 \\
            CE & \textsc{RoBERTa} & 32 & 7.5e-6 & 10 \\
            CE & \textsc{RoBERTa}+\textsc{mnli} & 32 & 7.5e-6 & 8 \\
            CE & \textsc{DeBERTa} & 16 & 1e-5 & 10 \\\midrule
            ART & \textsc{BERT} & 64 & 5e-6 & 9 \\
            ART & \textsc{RoBERTa} & 64 & 2.5e-6 & 5 \\
            ART & \textsc{DeBERTa} & 32 & 1e-6 & 9 \\\bottomrule

    \end{tabular}

    \normalsize
    \caption{Training hyperparameters (batch size, learning rate, epochs) for probed models.}

    \label{tbl:hyperparams}
\end{table}

\end{document}